%% file: main.tex
\documentclass[11pt]{videococo}

\usepackage{enumitem}
\usepackage{tabularx}
\usepackage{float}

\definecolor{top1orange}{RGB}{244,165,66}
\definecolor{top2orange}{RGB}{247,231,206}
\newcommand{\best}[1]{\cellcolor{top1orange}\textbf{#1}}
\newcommand{\second}[1]{\cellcolor{top2orange}#1}

\newfontfamily\arxivkeepbytesansmedium[Path=./]{ByteSans-Medium.ttf}
\newfontfamily\arxivkeepbytesansbold[Path=./]{ByteSans-Bold.ttf}
\newsavebox{\arxivdependencybox}
\AtBeginDocument{%
  \sbox{\arxivdependencybox}{%
    \includegraphics{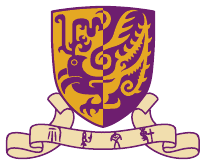}%
    \includegraphics{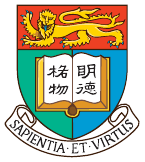}%
    \includegraphics{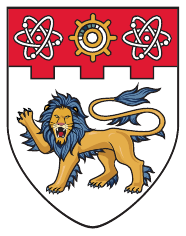}%
    \includegraphics{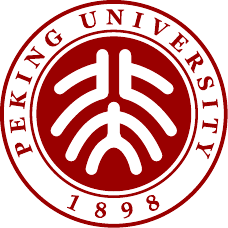}%
    \includegraphics{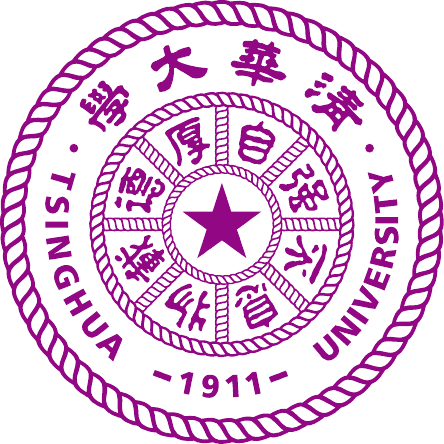}%
    \includegraphics{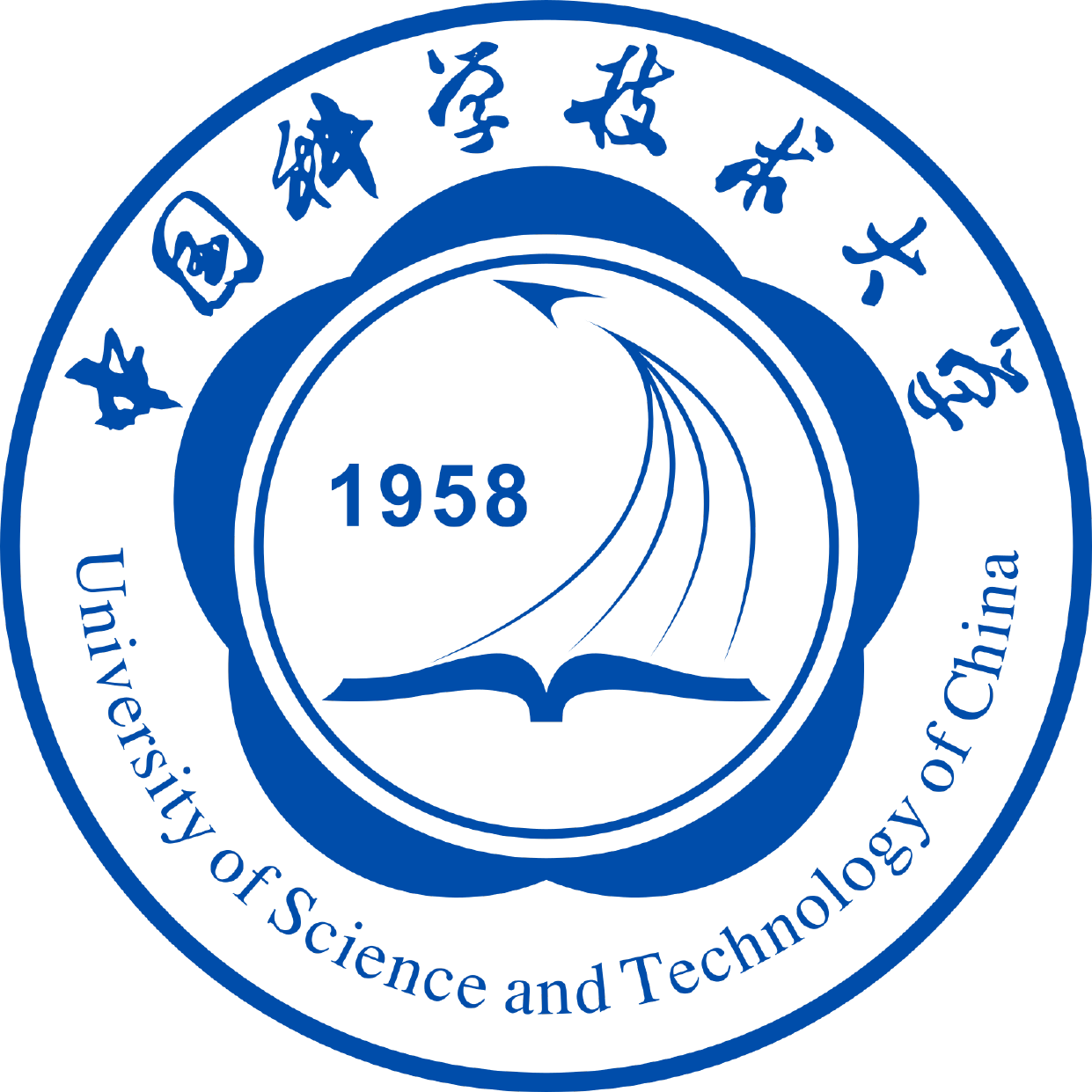}%
  }%
}

\title{\gradtext{VideoCoCo}: Code-as-CoT for Physically-Consistent Video Generation via an Agentic Dual-Engine System}

\author[1,*,\ddagger]{Haodong Li}
\author[2,*]{Tianfei Ren}
\author[2,*]{Xiaoxiao Ma}
\author[3,\dagger]{Chunmei Qing}
\author[2,\ddagger]{Zhen Fang}
\author[2]{\\Sipeng He}
\author[1]{Ziyu Guo}
\author[4]{Haoyu Wu}
\author[5]{Juanxi Tian}
\author[3]{Yihang Zou}
\author[6]{\\Ruichuan An}
\author[1]{Dongzhi Jiang}
\author[7]{Boxue Yang}
\author[8]{Ji Xie}
\author[6]{Xu Huang}
\author[9]{\\Wenhao Yan}
\author[10]{Jialv Zou}
\author[7]{Zhengrong Yue}
\author[11]{Yaxin Luo}
\author[6]{Xiaotong Li}
\author[9]{\\Yuzhu Wang}
\author[1]{Junyan Ye}
\author[1]{Jinjing Zhao}
\author[2]{Zehui Chen}
\author[2]{Lin Chen}
\author[6]{\\Renye Yan}
\author[2,\dagger]{Feng Zhao}
\author[1,\dagger]{Pheng-Ann Heng}

\affiliation[1]{CUHK}
\affiliation[2]{USTC}
\affiliation[3]{SCUT}
\affiliation[4]{HKU}
\affiliation[5]{NTU}
\affiliation[6]{PKU}
\affiliation[7]{SJTU}
\affiliation[8]{CMU}
\affiliation[9]{THU}
\affiliation[10]{HUST}
\affiliation[11]{MBZUAI}

\contribution[*]{Equal contribution.}
\contribution[\dagger]{Corresponding authors.}
\contribution[\ddagger]{Project leaders.}

\abstract{%
Text-to-video models have achieved remarkable visual quality, yet they still struggle to generate physically consistent dynamics because the temporal evolution of a scene must be inferred implicitly from a highly compressed text prompt. Existing chain-of-thought approaches introduce intermediate plans or visual states, but these representations are typically non-executable or temporally sparse, limiting their ability to instantiate and control the complete spatiotemporal process. To address this limitation, we introduce \textbf{VideoCoCo}, an agentic dual-engine framework in which executable Blender code serves as a process-level chain of thought. Given a text prompt, a coding agent synthesizes a Blender program that explicitly specifies the scene and its temporal evolution. The executable simulation engine runs the program to produce a deterministic spatiotemporal draft, which is subsequently transformed into a photorealistic video by a generative video engine through draft-conditioned editing. This decomposition separates process-level reasoning from high-fidelity visual realization. To adapt the video editor to simulated drafts, we construct \textbf{VideoCoCo-3K}, a curated dataset of \textbf{draft--instruction--target} triplets. VideoCoCo improves the OmniWeaving baseline from 0.475 to 0.558 on PhyGenBench and from 52.18 to 77.88 on VBench-2.0, achieving the best average score on both benchmarks. These results demonstrate that executable code provides an effective, controllable, and inspectable intermediate representation for physically consistent video generation.%
}

\date{August 2026}
\correspondence{\email{mickyhimself4@gmail.com}}
\checkdata[Project Page]{\url{https://github.com/micky-li-hd/VideoCoCo}}

\begin{document}
\maketitle

% ── Sections ─────────────────────────────────────
\input{section/intro}
\input{section/related}
\input{section/method}
\input{section/exp}
\input{section/conclusion}

% ── References ───────────────────────────────────
\bibliographystyle{unsrtnat}
\bibliography{aaai2027}

\end{document}

%% file: section/intro.tex
% ══════════════════════════════════════════════════
% 1. Introduction
% ══════════════════════════════════════════════════
\section{Introduction}
The pursuit of Artificial General Intelligence (AGI) requires machines to understand and model how the world evolves over time. Video generation provides a concrete means of representing such dynamics by transforming natural-language descriptions into temporally evolving visual sequences. Recent advances have enabled increasingly realistic and temporally coherent videos, positioning video generation as a promising pathway toward generative world modeling.

Realizing this potential, however, requires more than producing visually plausible videos. In the prevailing text-to-video paradigm, a model maps a text prompt to a complete visual sequence. Such prompts specify an event only at a highly compressed semantic level, leaving the physical principles governing its evolution largely unstated. The model must therefore recover a complete spatiotemporal process while simultaneously synthesizing its visual appearance. We refer to this mismatch between compressed intent and fully realized evolution as \textbf{Causal Opacity}. This raises a central question: \textit{\textbf{How can the process implicit in a prompt be externalized before final video generation?}}

\begin{figure}[t]
\centering
\includegraphics[width=\columnwidth]{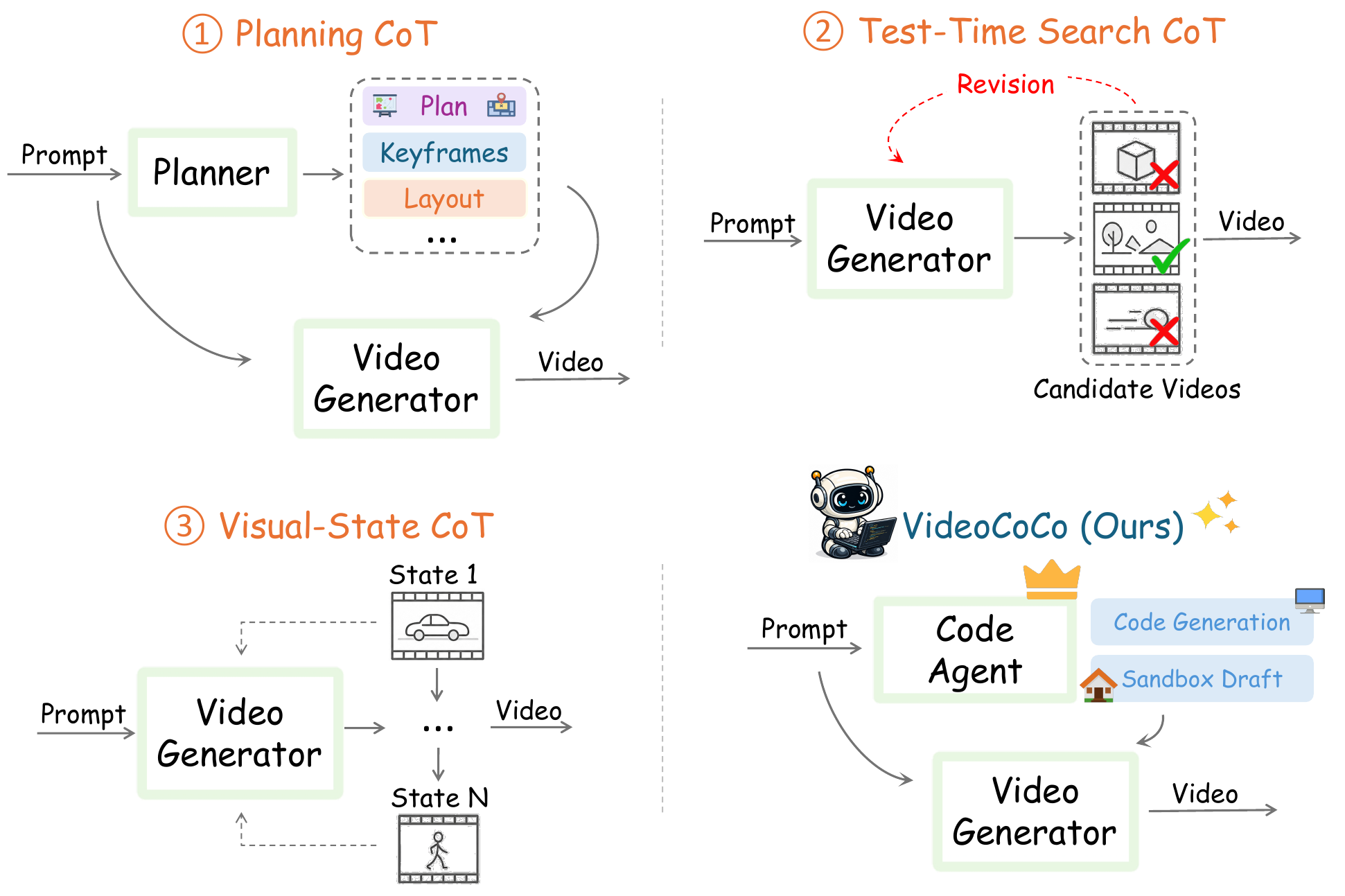}
\caption{Comparison of chain-of-thought paradigms for video generation. \textbf{(1) Planning CoT} externalizes reasoning as textual plans, keyframes, or layouts that condition a video generator. \textbf{(2) Test-Time Search CoT} samples multiple candidate videos and revises or selects among them at inference time. \textbf{(3) Visual-State CoT} reasons through a sequence of intermediate visual states within the generated video. In contrast, \textbf{VideoCoCo (Ours)} uses a code agent to synthesize executable code, renders a deterministic sandbox draft, and conditions the video generator on this draft, yielding a complete and inspectable process-level chain of thought rather than sparse or non-executable intermediates.}
\label{fig4}
\end{figure}

A natural way to address this question is to reason through an intermediate representation rather than directly in pixels. Recent work has begun to extend chain-of-thought reasoning to video generation: external-planning methods use text, layouts, or keyframes to guide synthesis~\citep{vchain2025,videodirectorgpt}; frame-chain methods reason through generated visual states~\citep{videorlvr,cheap2025}; and test-time methods search over or revise candidate trajectories~\citep{videot1,tbs2025}. Despite their different forms, these intermediates remain largely descriptive or selective, rather than instantiating a complete process that can be directly inspected and revised (Figure~\ref{fig4}). Recent advances in code agents point to a different possibility: executable code itself can serve as a process-level intermediate representation for video generation.

Building on this possibility, we introduce \textbf{VideoCoCo}, an agentic dual-engine framework in which executable code serves as the chain of thought. Given a text prompt, a coding agent synthesizes a Blender program that explicitly specifies the scene and its temporal evolution. The executable simulation engine runs the program in a sandboxed Blender environment to produce a deterministic spatiotemporal draft. Conditioned on this draft, the generative video engine realizes the simulated process as a photorealistic video through draft-conditioned editing. This division separates process-level reasoning from visual realization: the coding agent determines how the scene evolves, while the video editor realizes the instantiated process with high visual fidelity.

Realizing this dual-engine design requires supervision that connects simulated drafts with photorealistic targets while preserving their underlying spatiotemporal processes. To this end, we develop an agentic data-construction pipeline and construct \textbf{VideoCoCo-3K}, a dataset of draft--instruction--target triplets. These triplets adapt the generative video engine to realize simulated processes as photorealistic videos while retaining the spatiotemporal structure encoded in the drafts. Experiments on PhyGenBench and VBench-2.0 demonstrate that VideoCoCo substantially improves physical consistency over strong video generation baselines, validating executable code as an effective process-level intermediate representation.

In summary, our contributions are as follows:
\begin{itemize}
    \item We introduce \textbf{VideoCoCo}, an agentic dual-engine framework for physically consistent video generation. VideoCoCo uses executable Blender code as a process-level chain of thought, producing a spatiotemporal draft through simulation and realizing it as a photorealistic video through draft-conditioned editing.

    \item We construct \textbf{VideoCoCo-3K}, a dataset of \textbf{draft--instruction--target} triplets generated through an agentic data construction pipeline. The dataset provides aligned supervision for adapting the generative video engine to simulated drafts.

    \item VideoCoCo achieves state-of-the-art average performance on both PhyGenBench and VBench-2.0, substantially improving the OmniWeaving baseline, with particularly large gains in thermal and material dynamics.
\end{itemize}

%% file: section/related.tex
% ══════════════════════════════════════════════════
% 2. Related Work
% ══════════════════════════════════════════════════
\section{Related Work}

\subsection{Text-to-Video Diffusion Models}
Diffusion models have become the dominant paradigm for text-to-video (T2V) generation due to their strong visual quality and favorable scaling behavior. Early methods established the foundations of temporally coherent synthesis~\citep{ho2022video,singer2022make,ho2022imagen}. Building on these foundations, latent-diffusion approaches improved fidelity and efficiency through lightweight adaptation and tuning~\citep{blattmann2023align,chen2024videocrafter2,wang2025lavie,Wu_2023_ICCV}. More recently, large-scale video diffusion backbones have further strengthened temporal modeling, powering systems such as Open-Sora, HunyuanVideo, Wan, CogVideoX, and Step-Video-T2V~\citep{zheng2024open,kong2024hunyuanvideo,wan2025wan,yang2024cogvideox,ma2025stepvideot2vtechnicalreportpractice}. Beyond video synthesis itself, these models have also been explored as general visual and world representations~\citep{wang2026diffusion,tong2026scope,ye2026mind}. Complementary work strengthens generative world representations through reconstruction alignment, explicit coupling to 3D geometry, and scalable multi-agent, multi-view modeling~\citep{xie2025reconstruction,wu2026geometry,wu2026multiworld0}. Nevertheless, generating physically plausible and temporally coherent dynamics remains an open challenge~\citep{brooks2024video,meng2024towards,bansal2025videophy,zhang2025think,an2026genius,zhang2026well,li2026gebench,zheng2026pearl,yan2026proact,dai2026webvr,xu2026ad}.

\subsection{Physics-Aware Video Generation}
Existing approaches to physically plausible video generation can be broadly divided into explicit and implicit paradigms. Explicit methods directly incorporate physical knowledge through simulators, trajectory guidance, or structured constraints~\citep{wang2025wisa,xue2025phyt2v,yang2025vlipp,wang2025physctrl,zhang2025physchoreo}. Although effective for predefined phenomena, they often rely on fixed rules or templates that generalize poorly to open-world prompts. To reduce this dependence on hand-specified structure, implicit methods instead learn physical priors from data through preference optimization, reinforcement learning, or feature alignment~\citep{liu2024videodpo0,cai2025phygdpo,wang2025physcorr,li2025pisa,zhang2026physrvg}. Complementary work makes multimodal generative criteria explicit by deriving auto-rubrics as reward signals~\citep{tian2026auto}. This shift improves applicability, but offers limited control over prompt-specific processes because physical knowledge remains encoded indirectly through rewards, preferences, or learned representations. Despite their complementary strengths, neither paradigm typically grounds generation in an explicit spatiotemporal realization of the target process. VideoCoCo addresses this limitation by having an agent write and execute simulation code to produce a deterministic draft video, which subsequently guides a video-editing model toward a realistic result.

\subsection{Chain-of-Thought for Visual Generation}

\begin{figure*}[t]
\centering
\includegraphics[width=1\textwidth]{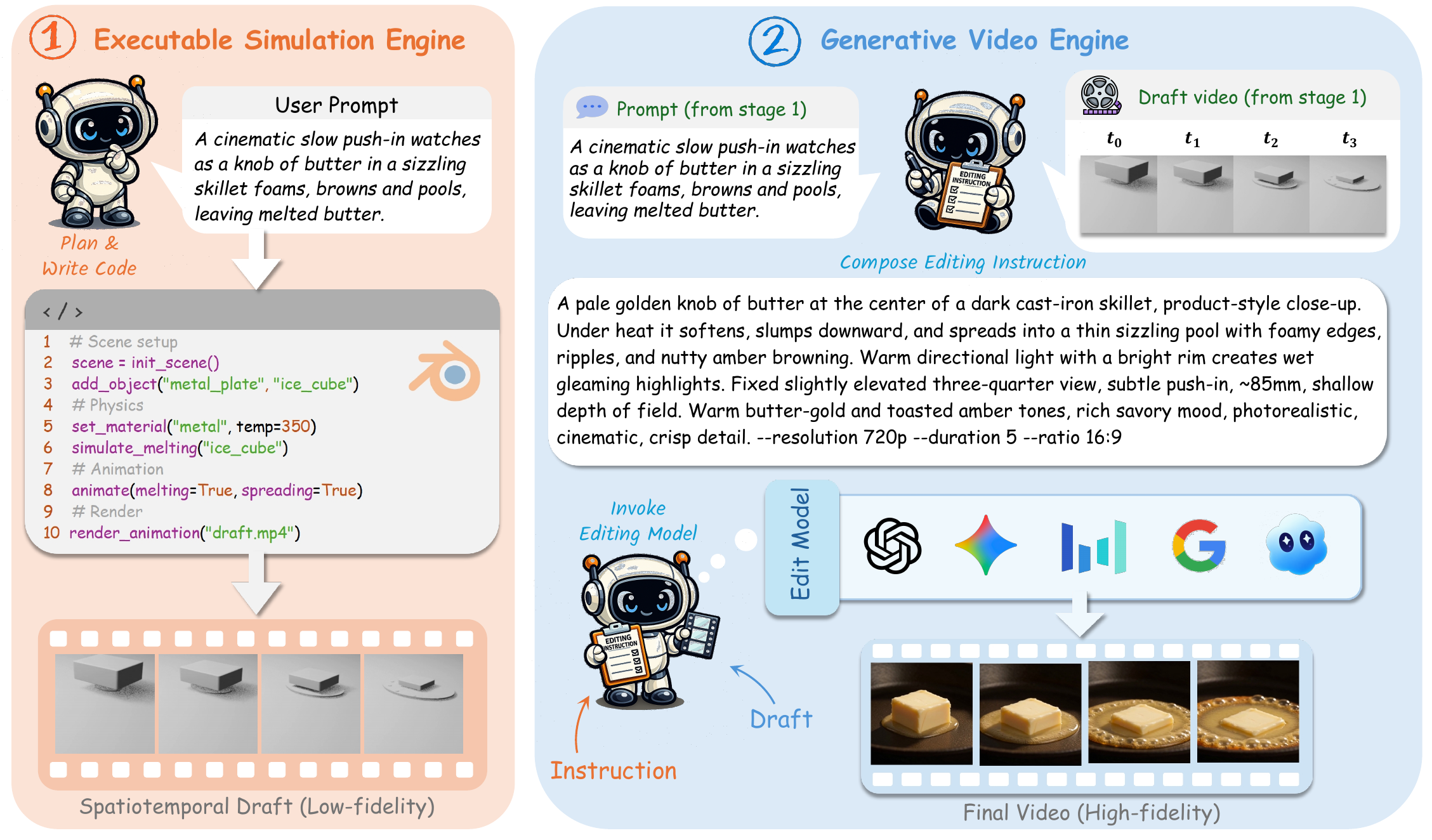}
\caption{Overview of the VideoCoCo dual-engine framework. \textbf{Stage 1 (Executable Simulation Engine):} given a user prompt, a coding agent plans and writes a self-contained Blender program that specifies the scene and its temporal evolution; executing this program in a sandbox renders a low-fidelity spatiotemporal draft. \textbf{Stage 2 (Generative Video Engine):} conditioned on the stage-1 prompt and draft video, an instruction agent composes an editing instruction, which together with the draft is passed to a video editing model to produce the final high-fidelity, physically consistent video. The executable draft supplies the process-level dynamics, while the editing model supplies photorealistic appearance.}
\label{fig2}
\end{figure*}

Chain-of-thought reasoning~\citep{wei2022cot} has recently been extended from language to visual generation. Existing video approaches generally follow three directions. One line of work externalizes reasoning through textual plans, layouts, or keyframes, as in VChain and VideoDirectorGPT~\citep{vchain2025,videodirectorgpt}. A second line embeds reasoning within the generated sequence itself, using intermediate visual states as in VideoRLVR and ChEaP~\citep{videorlvr,cheap2025}. Rather than prescribing intermediate states, a third line allocates additional computation at inference time to search over or revise generated trajectories, as in Video-T1 and temporal backtracking~\citep{videot1,tbs2025}. Beyond these video-generation paradigms, DraCo uses a draft as chain of thought for text-to-image preview and rare-concept generation~\citep{jiang2025draco}, while CoCo further represents the intermediate as executable code~\citep{li2026cococodecottexttoimage}. MetaPoint and VideoCoF introduce structured spatial control and temporal reasoning for agentic visual generation and unified video editing~\citep{zhou2026metapoint,Yang_2026_CVPR}. Although prior video CoT methods improve planning and structural consistency, their intermediate representations remain sparse, non-executable, or specialized to constrained environments. VideoCoCo instead extends executable Code-as-CoT from image preview to a complete temporal process: it renders a deterministic draft video that provides the downstream editor with dense spatiotemporal guidance over the target dynamics.

%% file: section/method.tex
% ══════════════════════════════════════════════════
% 3. Method
% ══════════════════════════════════════════════════
\section{Method}

We instantiate the Code-as-CoT idea as \textbf{VideoCoCo}, an agentic dual-engine framework that separates \emph{process-level reasoning} from \emph{visual realization}. As illustrated in Figure~\ref{fig2}, the two engines play complementary roles. An \emph{executable simulation engine} externalizes the process implicit in a text prompt as a runnable Blender program and renders it into a deterministic spatiotemporal draft, thereby committing the system to a concrete realization of the target dynamics before any pixels are synthesized. A \emph{generative video engine} then takes this draft as a structural condition and turns it into a photorealistic video, focusing its capacity on appearance rather than on reconstructing the process from scratch. This division is deliberate: it decouples the two aspects of physical-video generation that are hardest to satisfy simultaneously---faithful dynamics and photorealistic appearance---and lets each engine specialize in what it does best.

The remainder of this section follows this division. We first describe how the process is externalized as executable code, then how the resulting draft is realized as a photorealistic video, how we equip the video engine with the ability to read simulated drafts through a dedicated dataset, and finally how the editor is adapted and how the whole pipeline runs end-to-end at inference time.

\begin{figure*}[t]
\centering
\includegraphics[width=1\textwidth]{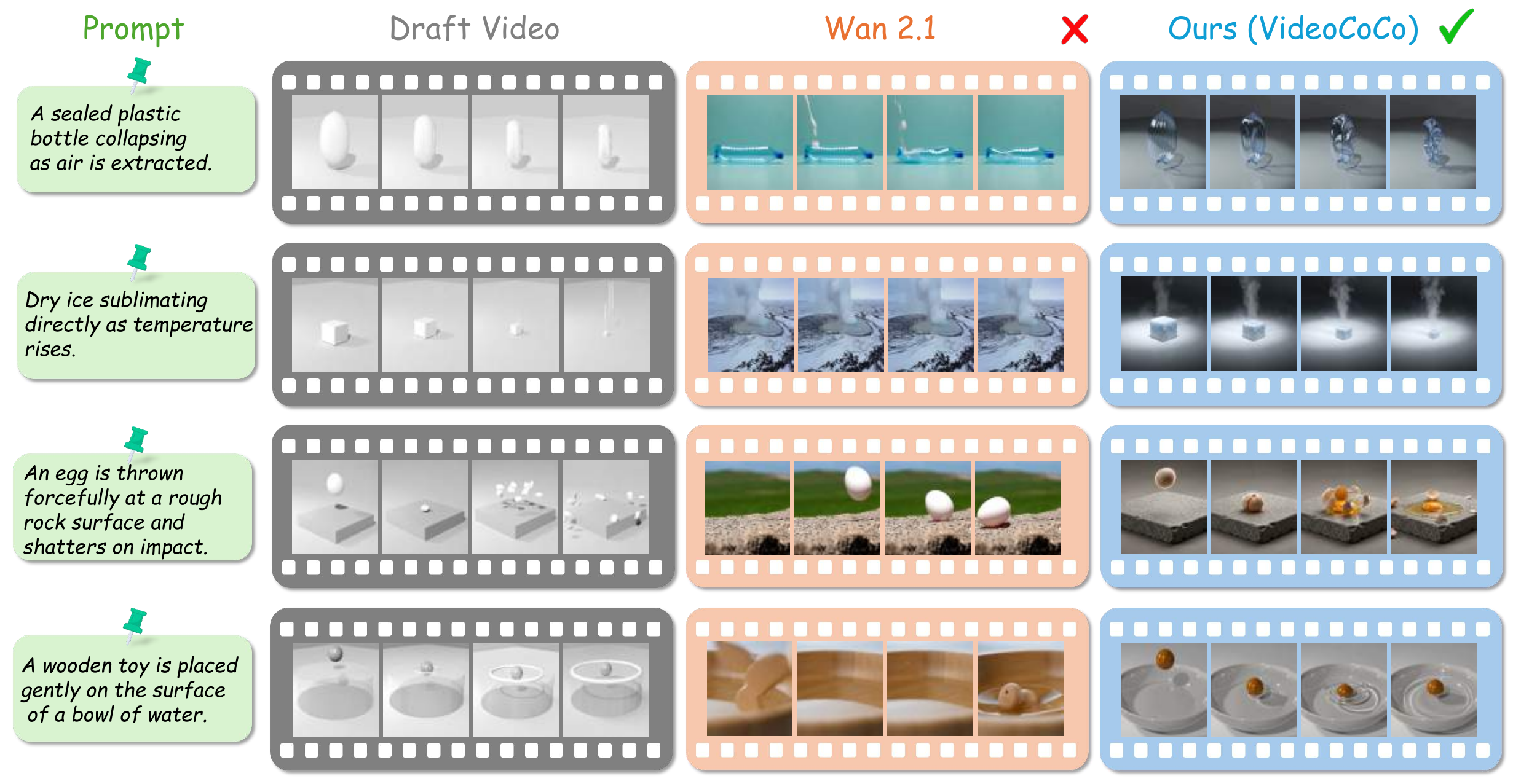}
\caption{\textbf{Qualitative comparison on representative physical processes.} OmniWeaving often produces visually plausible videos that violate the requested dynamics, whereas VideoCoCo follows the executable spatiotemporal draft and better preserves the intended processes, including sublimation, vacuum-induced collapse, impact shattering, and buoyancy.}
\label{fig3}
\end{figure*}

\subsection{Executable Simulation Engine}

A text prompt compresses a physical event into a few words, forcing a text-to-video model to reconstruct the entire spatiotemporal process from language alone. This burden is precisely where existing generators fail: appearance can be memorized from data, but a prompt-specific process rarely can. Instead of asking the video model to bear this burden implicitly, we let an agent write the process down explicitly. Our first engine therefore turns the prompt into an executable program and runs it in a sandbox, so that the target dynamics are instantiated \emph{before} any pixels are synthesized.

\paragraph{Code-as-CoT Program Synthesis.}
Given an input prompt $p$, a coding agent $A_{\mathrm{code}}$ synthesizes a self-contained Blender Python program
\begin{equation}
    c = A_{\mathrm{code}}(p),
\end{equation}
that specifies the scene, the objects and their physical properties, and the temporal evolution of the target event. We deliberately choose Blender Python as the representation of reasoning, rather than natural-language plans or sparse keyframes, for three reasons. First, it is \emph{explicit}: every object, motion, and interaction must be declared, leaving no room for the underspecification typical of textual plans. Second, it is \emph{executable}: the program commits the agent to a concrete process that can actually be run, not merely described. Third, it is \emph{inspectable}: the code can be read, edited, and re-executed, so the intermediate reasoning is transparent rather than hidden inside latent activations. Together, these properties mark the key departure from prior chain-of-thought representations for video, which remain either non-executable (text, layouts) or temporally sparse (isolated keyframes).

\paragraph{Sandboxed Execution and Draft Rendering.}
Code alone does not yet realize the process. We therefore execute the program in an isolated Blender environment $\mathcal{B}$ to obtain a rendered draft
\begin{equation}
    d = \mathcal{B}(c).
\end{equation}
Executing the program in a controlled runtime is essential: the sandbox provides a standardized set of primitives and enforces deterministic behavior, so that the same program always yields the same draft, and rendering errors are returned as diagnostic signals for code revision rather than remaining silent failures. The resulting draft is intentionally low-fidelity---a white-clay simulation without photorealistic materials or complex lighting---but it is \emph{temporally dense}: every frame corresponds to a physically instantiated state of the programmed process, from initial condition through intermediate stages to final outcome. In this sense $d$ is not a candidate output but a structural scaffold: it fixes what happens and when, leaving the question of how it should look to the next engine.

\subsection{Generative Video Engine}

The simulation draft solves the process, but not the appearance. Bridging this gap is the job of the second engine, whose role is deliberately narrow: rather than re-reasoning about the dynamics, it only translates the already instantiated process into a photorealistic video. This narrow scope is what allows the framework to be both physically consistent and visually faithful---the editor never has to guess what should happen, only how it should look.

\paragraph{Editing Instruction Construction.}
The original prompt $p$ and the rendered draft $d$ describe the target at very different levels of granularity. The prompt is compressed, abstract, and biased toward what the event \emph{is}, while the draft is dense, stylized, and biased toward how the event \emph{unfolds}. Passing $p$ directly to a video editor would therefore leave the appearance underspecified and risk generating motion that contradicts the draft. To reconcile the two, we introduce an instruction agent $A_{\mathrm{edit}}$ that reads both signals and composes an appearance-focused editing instruction:
\begin{equation}
    e = A_{\mathrm{edit}}(p, d).
\end{equation}
The instruction $e$ is written to describe the target subjects, materials, lighting, and cinematic style, and is explicitly discouraged from redefining the motion already present in $d$. It therefore complements, rather than competes with, the draft: $d$ dictates \emph{what happens}, while $e$ dictates \emph{what it looks like}.

\paragraph{Draft-Conditioned Video Editing.}
The two conditions are then fed jointly into a draft-conditioned video editor $G_{\theta}$, which produces the final video
\begin{equation}
    \hat{v} = G_{\theta}(d, e).
\end{equation}
The two conditions carry disjoint responsibilities: $d$ anchors the spatiotemporal structure of the target process, ensuring that the physical evolution is preserved, while $e$ specifies the photorealistic appearance the final video should exhibit. Under this decomposition, $G_{\theta}$ is asked to \emph{restyle} an already instantiated process rather than to imagine one from scratch---a substantially easier problem than direct text-to-video generation, and one that closely aligns with the strengths of modern video-editing models (see Figure~\ref{fig3} for a qualitative illustration). However, the draft follows a distinctive white-clay simulation style that lies outside the training distribution of off-the-shelf editors, which either ignore its motion or hallucinate appearance that conflicts with it. Turning $G_{\theta}$ into a competent draft-conditioned editor therefore requires targeted supervision, which motivates the dataset introduced next.

\subsection{VideoCoCo-3K: Data for Draft-Conditioned Editing}
\label{sec:data_construction}
The core difficulty in adapting $G_{\theta}$ is that publicly available video-editing datasets contain only natural-video pairs. There is essentially no supervision that pairs a simulated white-clay draft with a photorealistic video sharing the \emph{same} spatiotemporal process, which is precisely what our second engine needs to learn from. We therefore construct \textbf{VideoCoCo-3K}, a dataset of \textbf{draft--instruction--target} triplets that provides exactly this alignment.

\paragraph{Teacher-Based Triplet Construction.}
We build the dataset by running the pipeline of Sec.~3.1--3.2 at scale. For each prompt $p_i$ collected from our source distribution, the executable simulation engine produces a rendered draft $d_i$ and the instruction agent composes an editing instruction $e_i$. To obtain the corresponding photorealistic target, we invoke a high-fidelity teacher editor $G_{\mathrm{T}}$ once per triplet:
\begin{equation}
    y_i = G_{\mathrm{T}}(d_i, e_i).
\end{equation}
We instantiate $G_{\mathrm{T}}$ as Seedance~2.0, chosen because it is strong enough to produce photorealistic frames while faithfully preserving the motion carried by $d_i$---the two properties that a training target for draft-conditioned editing must simultaneously satisfy. The resulting dataset is
\begin{equation}
    \mathcal{D}_{\mathrm{VideoCoCo\text{-}3K}} = \left\{(d_i, e_i, y_i)\right\}_{i=1}^{3000},
\end{equation}
in which each triplet aligns a physically grounded process with its photorealistic realization. We exclude all prompts from the evaluation benchmarks, along with their near-duplicates. We additionally retain each original prompt and Blender program as metadata, allowing the dataset to be inspected, extended, or regenerated as the pipeline evolves.

\subsection{Editor Adaptation and Inference}

With the aligned triplets in place, the remaining question is how to teach $G_{\theta}$ to consume simulated drafts and how the whole pipeline runs at inference time from a text prompt alone.

\paragraph{Training Objective.}
We adapt the draft-conditioned editor $G_{\theta}$ on the VideoCoCo-3K triplets $(d_i, e_i, y_i)$, where the draft and instruction serve as conditions and the photorealistic target provides supervision. Let $z_0$ denote the latent of the target $y$ and $z_t$ its noised version at diffusion timestep $t$. The editor is trained with a standard conditional denoising objective
\begin{equation}
    \mathcal{L}(\theta) = \mathbb{E}_{(d,e,y),\,t,\,\epsilon}\!\left[\left\|\epsilon - \epsilon_{\theta}(z_t, t, d, e)\right\|_2^2\right],
\end{equation}
with $\epsilon \sim \mathcal{N}(0, I)$. Minimizing this objective teaches the editor to synthesize a photorealistic video whose noise structure is consistent with the target while remaining conditioned on both the simulated draft and the editing instruction. We initialize $G_{\theta}$ from the OmniWeaving base generator and consider two adaptation strategies---full fine-tuning and parameter-efficient LoRA---whose empirical comparison is deferred to Sec.~4.3.

\paragraph{End-to-End Inference.}
At inference time, VideoCoCo operates end-to-end from a text prompt alone. Given a prompt $p$, the coding agent synthesizes a Blender program $c$, the sandbox renders a deterministic draft $d$, the instruction agent composes an editing instruction $e$, and the adapted editor produces the final video $\hat{v}$. No ground-truth draft, human annotation, or auxiliary input is required: the user supplies only the prompt, and the executable draft is generated automatically as the process-level chain of thought. Because every intermediate artifact---the program, the draft, and the instruction---is either code or a rendered video, the entire pipeline is \emph{fully automatic, inspectable, and reproducible}, which we regard as a defining property of Code-as-CoT for video generation.

\input{tables/main1}
\input{tables/main2}
\input{tables/main3}

%% file: tables/main1.tex
\begin{table*}[t]
\centering
\setlength{\tabcolsep}{5pt}
\renewcommand{\arraystretch}{1.15}

\begin{tabular*}{\textwidth}{@{\extracolsep{\fill}}clccccc}
\toprule
& Method
& Mechanics ($\uparrow$)
& Optics ($\uparrow$)
& Thermal ($\uparrow$)
& Material ($\uparrow$)
& Average ($\uparrow$) \\
\midrule

\multirow{3}{*}{\rotatebox{90}{Closed}}
& Pika~\citep{pika}
& 0.35
& 0.56
& 0.43
& 0.39
& 0.44 \\

& Gen-3~\citep{gen3}
& 0.45
& 0.57
& 0.49
& \second{0.51}
& 0.51 \\

& Kling~\citep{kling}
& 0.45
& \second{0.58}
& 0.50
& 0.40
& 0.49 \\

\midrule

\multirow{8}{*}{\rotatebox{90}{Open}}
& CogVideoX~\citep{yang2024cogvideox}
& 0.39
& 0.55
& 0.40
& 0.42
& 0.45 \\

& Open-Sora V1.2~\citep{zheng2024open}
& 0.43
& 0.50
& 0.44
& 0.37
& 0.44 \\

& LaVie~\citep{wang2025lavie}
& 0.30
& 0.44
& 0.38
& 0.32
& 0.36 \\

& Vchitect-2.0~\citep{fan2025vchitect}
& 0.41
& 0.56
& 0.44
& 0.37
& 0.45 \\

& HunyuanVideo~\citep{kong2024hunyuanvideo}
& 0.33
& 0.39
& 0.26
& 0.30
& 0.33 \\

& Wan2.2-TI2V-5B~\citep{wan22}
& \second{0.55}
& \second{0.58}
& \best{0.53}
& 0.50
& \second{0.54} \\

& Cosmos-Predict2.5~\citep{cosmos}
& 0.30
& 0.33
& 0.41
& 0.39
& 0.35 \\

& LTX-Video-2B~\citep{ltxvideo}
& 0.51
& \second{0.58}
& 0.48
& 0.45
& 0.51 \\

\midrule

\multirow{2}{*}{\rotatebox{90}{Ours}}
& OmniWeaving~\citep{omniweaving}
& 0.48
& 0.56
& 0.43
& 0.39
& 0.48 \\

& \;+ VideoCoCo
& \best{0.56}
& \best{0.61}
& \second{0.51}
& \best{0.53}
& \best{0.56} \\

\bottomrule
\end{tabular*}

\caption{
Physical-consistency comparison on PhyGenBench across closed- and
open-source video generators. We report per-category consistency scores
in $[0,1]$ for mechanics, optics, thermal, and material dynamics,
together with their average; higher is better. Adding VideoCoCo to the
OmniWeaving base generator lifts every category and raises the average
from 0.48 to 0.56, the best overall result, with the largest gains on
material and thermal dynamics, where appearance-driven priors are
weakest. \colorbox{top1orange}{\textbf{Orange}} and
\colorbox{top2orange}{Champagne} cells denote the best (top-1) and
second-best (top-2) results per column.
}
\label{tab:phygenbench}
\end{table*}

%% file: tables/main2.tex
  \begin{table*}[h]
  \centering
  \setlength{\tabcolsep}{6pt}
  \renewcommand{\arraystretch}{1.15}

  % 删除列格式末尾的 @{}
  \begin{tabular*}{\textwidth}{@{\extracolsep{\fill}}clcccc}
  \toprule
  & Method
  & Mechanics ($\uparrow$)
  & Thermotics ($\uparrow$)
  & Material ($\uparrow$)
  & Average ($\uparrow$) \\
  \midrule

  \multirow{2}{*}{\rotatebox{90}{Closed}}
  & Sora~\citep{brooks2024video}
  & 62.22\%
  & 43.36\%
  & 64.94\%
  & 56.84\% \\

  & Kling 1.6~\citep{kling16}
  & 65.55\%
  & 59.46\%
  & 68.00\%
  & 64.34\% \\[4pt]

  \midrule

  \multirow{2}{*}{\rotatebox{90}{Open}}
  & HunyuanVideo~\citep{kong2024hunyuanvideo}
  & 76.09\%
  & 56.52\%
  & 64.37\%
  & 65.66\% \\

  & CogVideoX-1.5~\citep{yang2024cogvideox}
  & \second{80.80\%}
  & \second{67.13\%}
  & \best{83.19\%}
  & \second{77.04\%} \\

  \midrule

  \multirow{2}{*}{\rotatebox{90}{Ours}}
  & OmniWeaving~\citep{omniweaving}
  & 62.79\%
  & 52.08\%
  & 41.67\%
  & 52.18\% \\

  & \;+ VideoCoCo
  & \best{92.31\%}
  & \best{72.92\%}
  & \second{68.42\%}
  & \best{77.88\%} \\

  \bottomrule
  \end{tabular*}

  \caption{
  Physical-plausibility comparison on VBench-2.0, reported as
  per-dimension plausibility percentages over mechanics, thermotics, and
  material, with their average; higher is better. Coupling VideoCoCo with
  OmniWeaving improves the average from 52.18\% to 77.88\% (a gain of
  25.70 points) and attains the best mechanics and thermotics scores among
  all systems, trailing only CogVideoX-1.5 on material.
  \colorbox{top1orange}{\textbf{Orange}} and
  \colorbox{top2orange}{Champagne} cells denote the best (top-1) and
  second-best (top-2) results per column.
  }
  \label{tab:vbench2_physics}
  \end{table*}

%% file: tables/main3.tex
  \begin{table}[t]
  \centering
  \small

  \setlength{\tabcolsep}{3pt}
  \setlength{\extrarowheight}{1pt}
  \renewcommand{\arraystretch}{1.4}

  \begin{tabular}{lccccc}
  \toprule
  Method
  & Mech.\,($\uparrow$)
  & Opt.\,($\uparrow$)
  & Therm.\,($\uparrow$)
  & Mat.\,($\uparrow$)
  & Avg.\,($\uparrow$) \\
  \midrule

  OmniWeaving~\citep{omniweaving}
  & 0.48
  & 0.56
  & 0.43
  & 0.39
  & 0.48 \\

  \midrule

  + VideoCoCo (Tune-Free)
  & 0.50
  & 0.53
  & 0.48
  & \underline{0.51}
  & 0.51 \\

  + VideoCoCo (Full-Tune)
  & \underline{0.51}
  & \textbf{0.61}
  & \textbf{0.51}
  & 0.49
  & \underline{0.54} \\

  + VideoCoCo (LoRA-Tune)
  & \textbf{0.56}
  & \textbf{0.61}
  & \textbf{0.51}
  & \textbf{0.53}
  & \textbf{0.56} \\

  \bottomrule
  \end{tabular}

  \caption{
  Ablation of video-editor adaptation strategies on PhyGenBench.
  All variants share the same executable draft-generation pipeline and
  differ only in how the editor is adapted: Tuning-Free, Full-Tune, or
  LoRA-Tune. The Tuning-Free variant already surpasses OmniWeaving
  (0.48 $\rightarrow$ 0.51), isolating the contribution of executable
  drafting, while LoRA-Tune yields the best average (0.56), showing
  that drafting and editor adaptation are complementary.
  \textbf{Bold} and \underline{underline} denote the best (top-1) and
  second-best (top-2) results per column.
  }
  \label{tab:ablation_phygenbench}
  \end{table}

%% file: section/exp.tex
% ======================================================
% 4. Experiments
% ======================================================
\section{Experiments}

This section shows that Code-as-CoT drafting improves the physical consistency of generated video, achieving the best average score on both PhyGenBench and VBench-2.0, and that both the executable draft and the editing stage contribute to the gain.

\subsection{Experimental Setup}
\textbf{Datasets.} We evaluate on two physics-focused video benchmarks. \textbf{PhyGenBench}~\citep{phygenbench} evaluates physical commonsense across four categories: mechanics, optics, thermal, and material dynamics. We follow its official evaluation protocol, which uses GPT-4o~\citep{gpt4o} as the MLLM judge to assess how faithfully a generated video adheres to the physical principles implied by its prompt, providing a comprehensive measure of the physical knowledge instantiated by a generative model. \textbf{VBench-2.0}~\citep{vbench2} targets intrinsic faithfulness beyond surface quality; we adopt its per-dimension protocol on a pre-registered subset that balances temporal and perceptual/semantic factors. We focus on its physical dimensions---Mechanics ($\uparrow$; e.g., gravity, buoyancy, stress), Thermotics ($\uparrow$; e.g., vaporization, freezing), and Material ($\uparrow$; e.g., color mixing, solubility)---each probed by a dedicated prompt suite and scored to reflect adherence to real-world physical laws.

\textbf{Baselines.} We group baselines by access. Closed-source, API-only systems include Pika~\citep{pika}, Gen-3~\citep{gen3}, Kling~\citep{kling}, and Sora~\citep{brooks2024video}. Open-source systems with publicly released weights include CogVideoX~\citep{yang2024cogvideox}, Open-Sora~\citep{zheng2024open}, LaVie~\citep{wang2025lavie}, Vchitect-2.0~\citep{fan2025vchitect}, HunyuanVideo~\citep{kong2024hunyuanvideo}, Wan2.2-TI2V-5B~\citep{wan22}, Cosmos-Predict2.5~\citep{cosmos}, and LTX-Video~\citep{ltxvideo}. VideoCoCo augments the OmniWeaving~\citep{omniweaving} base generator, which we also report on its own to isolate our contribution.

\textbf{Metrics.} PhyGenBench reports a per-category consistency score in $[0,1]$ and their average. VBench-2.0 reports per-dimension plausibility as a percentage. Higher is better on every metric.

\textbf{Implementation.} VideoCoCo writes Blender programs, renders deterministic draft videos in a sandbox, and refines them with a video editor trained on VideoCoCo-3K. We evaluate three editor regimes: tuning-free, LoRA, and full tuning.

\subsection{Code-as-CoT Improves Physical Consistency}
\textbf{PhyGenBench.} Table~\ref{tab:phygenbench} compares VideoCoCo against a broad set of closed- and open-source generators, including recent strong open models such as Wan2.2-TI2V-5B~\citep{wan22}, Cosmos-Predict2.5~\citep{cosmos}, and LTX-Video~\citep{ltxvideo}. Adding VideoCoCo to OmniWeaving raises the average consistency score from 0.475 to 0.558, the best overall result in the table, ahead of the strongest open baseline Wan2.2-TI2V-5B at 0.544. VideoCoCo attains the best per-category score on mechanics (0.558), optics (0.613), and material (0.525), and ranks a close second on thermal (0.511), where Wan2.2-TI2V-5B leads narrowly at 0.533. The largest gains over the OmniWeaving base appear on material (+0.133) and thermal (+0.078), precisely the categories where appearance-driven generators, lacking an explicit dynamics reference, are weakest.

\textbf{Takeaway.} Executable drafting helps most where visual priors are weakest, which confirms that code contributes physical dynamics rather than surface realism.

\textbf{VBench-2.0.} Table~\ref{tab:vbench2_physics} reports physical plausibility on VBench-2.0. OmniWeaving alone scores 52.18\% on average, and adding VideoCoCo raises the average to 77.88\%, a gain of 25.70 points over the base model and the best average among all systems. VideoCoCo attains the best mechanics (92.31\%) and thermotics (72.92\%) scores, improving the base generator by 29.52 and 20.84 points respectively, and ranks second on material (68.42\%), trailing only CogVideoX-1.5 (83.19\%).

\textbf{Takeaway.} The VBench-2.0 pattern matches PhyGenBench: simulation-grounded drafts lift the dimensions that data-driven priors handle worst.

\subsection{Both Drafting and Tuning Contribute}
Table~\ref{tab:ablation_phygenbench} ablates the editor adaptation regime on PhyGenBench, holding the executable draft-generation pipeline fixed so that only the way the video editor is adapted varies. The \textit{tuning-free} variant, which conditions the unmodified OmniWeaving editor directly on the simulation draft, already improves the average from 0.475 to 0.506 and lifts the thermal and material categories most (0.433$\rightarrow$0.478 and 0.392$\rightarrow$0.508). Since no editor parameters are updated, this gain is attributable entirely to the executable draft, confirming that the draft---not additional training---supplies the physical dynamics. Adapting the editor on VideoCoCo-3K adds a further, complementary gain: \textit{full tuning} raises the average to 0.535, and \textit{LoRA tuning} attains the best result, 0.558, improving every category over the base generator.

Notably, LoRA outperforms full fine-tuning (0.558 vs.\ 0.535) despite updating far fewer parameters. We attribute this to the narrow nature of the adaptation: the editor does not need to relearn general video priors, but only to acquire the transferable skill of restyling a white-clay simulation draft into a photorealistic video while preserving its motion. Restricting adaptation to a low-rank subspace retains the strong visual priors of the base generator and learns this draft-to-realistic mapping, whereas full fine-tuning is more prone to overfitting the limited triplet data and drifting away from those priors.

\textbf{Takeaway.} Each stage contributes. Drafting supplies most of the physical gain, and editor tuning refines appearance while preserving the simulated dynamics; a lightweight LoRA adaptation is sufficient---and preferable---to full fine-tuning.

\textbf{Qualitative results.} Figure~\ref{fig3} illustrates these gains: on prompts such as dry-ice sublimation and vacuum-induced bottle collapse, the OmniWeaving baseline produces plausible frames that violate the intended process, whereas VideoCoCo follows the executable draft and realizes the correct dynamics with photorealistic appearance.

%% file: section/conclusion.tex
% ══════════════════════════════════════════════════
% 5. Conclusion
% ══════════════════════════════════════════════════
\section{Conclusion}

In this paper, we introduced VideoCoCo, an agentic dual-engine framework that uses executable code as a process-level chain of thought to enforce physical consistency in text-to-video generation. By decoupling spatiotemporal simulation from photorealistic editing, VideoCoCo achieves state-of-the-art physical plausibility on PhyGenBench and VBench-2.0. However, our approach introduces additional inference latency and is bounded by the expressiveness of the underlying Blender simulator, making highly complex phenomena like turbulent fluids challenging to synthesize zero-shot. Future work will explore integrating specialized physical engines (e.g., Taichi) and investigate knowledge distillation to internalize these executable priors directly into end-to-end video models, ultimately eliminating the need for inference-time simulation.